\documentclass[10pt, a4paper]{article}
\usepackage{lrec}
\usepackage{multibib}
\newcites{languageresource}{Language Resources}
\usepackage{graphicx}
\usepackage{tabularx}
\usepackage{soul}
\usepackage{titlesec}
\titleformat{\section}{\normalfont\large\bf\center}{\thesection.}{1em}{}
\titleformat{\subsection}{\normalfont\SmallTitleFont\bf\raggedright}{\thesubsection.}{1em}{}
\titleformat{\subsubsection}{\normalfont\normalsize\bf\raggedright}{\thesubsubsection.}{1em}{}
\renewcommand\thesection{\arabic{section}}
\renewcommand\thesubsection{\thesection.\arabic{subsection}}
\renewcommand\thesubsubsection{\thesubsection.\arabic{subsubsection}}
\usepackage{epstopdf}
\usepackage[utf8]{inputenc}

\usepackage{hyperref}
\usepackage{xstring}
\usepackage{todonotes}
\usepackage{color}

\title{A Cross-Genre Ensemble Approach to Robust Reddit Part of Speech Tagging}

\name{Shabnam Behzad, Amir Zeldes}

\address{Corpling Lab \\
    Georgetown University \\
    shabnam@cs.georgetown.edu, 
    amir.zeldes@georgetown.edu\\
    }

\abstract{
Part of speech tagging is a fundamental NLP task often regarded as solved for high-resource languages such as English. Current state-of-the-art models have achieved high accuracy, especially on the news domain. However, when these models are applied to other corpora with different genres, and especially user-generated data from the Web, we see substantial drops in performance. In this work, we study how a state-of-the-art tagging model trained on different genres performs on Web content from unfiltered Reddit forum discussions. More specifically, we use data from multiple sources: OntoNotes, a large benchmark corpus with `well-edited' text, the English Web Treebank with 5 Web genres, and GUM, with 7 further genres other than Reddit. We report the results when training on different splits of the data, tested on Reddit. Our results show that even small amounts of in-domain data can outperform the contribution of data an order of magnitude larger coming from other Web domains. To make progress on out-of-domain tagging, we also evaluate an ensemble approach using multiple single-genre taggers as input features to a meta-classifier. We present state of the art performance on tagging Reddit data, as well as error analysis of the results of these models, and offer a typology of the most common error types among them, broken down by training corpus. \\ \newline  }

\begin{document}

\maketitleabstract

\section{Introduction}

With the rapid growth of social media platforms and general public participation on the Internet, user-generated content has become one of the main data resources for different applications \cite{SanguinettiBoscoCassidyEtAl2020}. Textual data from these platforms are being used in many NLP tasks even though they are often not well-structured, and deviate from prescriptive language norms. The combination of such heterogeneous data and differences with typical kinds of training data (often newswire language) are hence challenging to work with: different types of noise are introduced into these datasets because of non-standard lexical items, spelling inconsistencies, informal abbreviations, and linguistic errors~\cite{gui-etal-2017-part,meftah-semmar-2018-neural}. 

Part of speech tagging is a fundamental NLP task which has long been studied, and, based on standard benchmarks, now seems nearly solved: for example, recent approaches have reached an accuracy of 97.85\%~\cite{akbik-etal-2018-contextual} on the Wall Street Journal corpus, essentially approaching human levels of accuracy. However, when state-of-the-art models are evaluated on out of domain data, we observe a drop in performance~\cite{derczynski-etal-2013-twitter}; this could be the result of differences in topic, writing style and epoch between training and testing data~\cite{manning2011part}. At the same time, high quality POS tagging is particularly pertinent for non-standard language, since exposing parts of speech in unusual text types gives access to underlying categories (e.g. proper nouns, predicates) which are difficult to recognize on a textual basis when they have unusual forms. Because the resulting POS tags are frequently used as part of downstream NLP tasks, errors caused by the tagger can propagate and affect the results of these downstream tasks as well~\cite{foster2011hardtoparse}. Thus, NLP tasks can benefit substantially from high accuracy, domain-robust POS tagging.

Enhancing the performance of taggers for social media data in particular has been studied before. Most of these studies, however, have focused on data from Twitter, which diverges from standard language strongly, but also represents a very narrow subdomain of user-generated content. In this work, we focus on a different platform, Reddit, which has a more heterogeneous text structure from Twitter. We compare the performance of the state-of-the-art tagging framework Flair~\cite{akbik-etal-2018-contextual} trained on different genres and tested on Reddit data, and provide a deep analysis of the errors produced in each of the models. Our initial results suggest that even small amounts of in-domain data used in training can outperform the contribution of data an order of magnitude larger but from other domains, despite the fact that most of the data sources used in this paper come from a range of Web genres themselves. In order to achieve progress on generalization to new domains, we also evaluate an ensemble model which uses the predictions of multiple models trained on different genres as features. We observe the effectiveness of these features by an ablation study and report the results.

\section{Related Work}
Over the past decades there has been a growing body of work focusing on POS tagging and domain adaptation. Many approaches have been proposed to improve tagging performance using different models such as Conditional Random Fields, Hidden/Maximum Entropy Markov
Models, linear classifiers and neural architectures~\cite{mueller-etal-2013-efficient,NIPS2014_5563,huang2015bidirectional,choi-2016-dynamic,qi2018universal,akbik-etal-2018-contextual}.

With the growth of social media and the tremendous amount of user-generated textual data available, researchers now analyze and use these data in many different NLP tasks~\cite{liu-etal-2018-parsing}. Studies show that the performance of NLP tools including POS taggers typically degrades when the models are tested on unedited text such as tweets~\cite{ritter-etal-2011-named}, however, retraining the models on in-domain data can improve performance~\cite{neunerdt2013part}.~\newcite{Evert2009IsPT} presented an evaluation of various POS taggers in German when trained on newspaper corpora and then tested on less standardized
text genres such as Web corpora and observed a drop in performance. They also analyzed how tagging different web genres could present different levels of difficulty for the trained models. More specifically, they found TV episode guide, online forum, conference information and news report data to be harder than other text genres.

Studies of tagging specifically for the heterogeneous space of Reddit text remain outstanding. Previous research has studied the problem of POS tagging on social media data primarily by targeting Twitter. Some have proposed new tagging schemes and released new annotated datasets.~\newcite{ritter-etal-2011-named} added new tags for Twitter specific phenomena such as \#hashtags and @usernames.~\newcite{gimpel-etal-2011-part}, developed a POS tagset specifically for English Twitter and a new dataset of manually tagged tweets.~\newcite{owoputi-etal-2013-improved} released a new manually annotated dataset for English Twitter POS tagging along with a part of speech tagger for online conversational text. There have also been efforts on POS tagging for other languages such as Irish~\cite{lynn-etal-2015-minority} and Italian~\cite{bosco2016overview}. A shared task on the Automatic Linguistic Annotation of Computer-Mediated Communication (CMC) and Web Corpora for German was also organized by~\newcite{beisswenger-etal-2016-empirist} to observe whether both CMC and Web corpora can be processed using the same methodologies and whether improved models can be introduced for tokenization and POS tagging of German computer-mediated communication using the new annotated data and other techniques, such as domain adaptation. Domain adaptation and regularization are helpful techniques when dealing with low-resource text types and many studies have focused on enhancing POS tagging using such methods~\cite{meftah-etal-2019-joint,marz-etal-2019-domain}.

Some studies have conducted error analysis of social media taggers, though not yet on Reddit.~\newcite{derczynski-etal-2013-twitter} evaluated the performance of existing POS taggers on Twitter datasets. They also provide an in-depth analysis of the errors on the tokens that were not seen during training. They report gold standard error, slang, genre-specific tokens and unseen proper nouns among the common error categories. ~\newcite{albogamy-ramsay-2016-fast} also evaluate state-of-the-art POS taggers on Arabic tweets. They categorize errors into 2 groups: errors on Arabic words and errors on non-Arabic tokens. Each of these groups includes subcategories such as named entities that were not seen during training, concatenation of multiple words, emoticons, foreign words, and others.

To the best of our knowledge, such in-depth studies have not yet been done on Reddit even though it is widely used as a data source for different NLP tasks. In this paper, we study genre effects on POS tagging accuracy for Reddit text when training data itself comes entirely from the Web (but not from Reddit), from other large benchmark resources such as OntoNotes \cite{HovyMarcusPalmerEtAl2006}) or both. We provide a detailed error analysis of different models, which suggests that some of the difficulties in tagging Reddit are not only due to the noisy nature of text online, but also to specific language use in Reddit as a genre. We also present an ensemble tagging approach that has a higher accuracy than the best single training genre baseline.

\section{Approach}\label{app}
\subsection{Data}

In this study, we use three different corpora with different genres. The main corpus we used is GUM (the Georgetown University Multilayer corpus~\citelanguageresource{GUM}) which was chosen because it contains gold standard tagged Reddit data. The corpus has manual annotation for different tasks such as POS tagging, lemmatization, dependency parses, discourse parses and entity and coreference resolution~\cite{Zeldes2017}, though the latter layers are not used in this study. Currently, GUM comprises about 130,000 tokens with data from 8 different genres in English, which, aside from Reddit, include creative commons licensed Academic papers and Fiction, Biographies (Bio) from Wikipedia, WikiNews Interviews and News stories, Wikivoyage travel guides and Wikihow how-to guides (Whow). Importantly, all of the data in the corpus was harvested from the Web, meaning that even when training on other genres and testing on Reddit, only data which is encountered on the Internet is involved. In order to get comparable numbers for models trained on other popular benchmark resources, we also use larger corpora such as EWT~\citelanguageresource{EWT} (about 250,000 tokens of data from the Web) and English OntoNotes~\citelanguageresource{OntoNotes} (about 2.6 million tokens, mostly from edited print texts and spoken data) in our experiments which are mainly used for POS tagging evaluations.

We have 12 different training splits for this task; for every GUM genre except for Reddit, we use all available data as the training set. Reddit has the smallest training set since we need some of the documents for development and test sets. Out of 11,182 annotated Reddit tokens in GUM version 5, we use 5,727 tokens for training, and 2,489 tokens for development and 2,966 tokens for the test set. The Reddit documents are from different discussion threads which makes evaluation more realistic.

We also create a split that contains the training data of multiple genres (Reddit, Academic, Bio, Fiction, Interview, News, Voyage, Whow) (\textit{Multiple Genres}) and another one which contains training data from the same genres except for Reddit (\textit{Multiple Genres w/o Reddit}). The size of all these training sets is shown in Table~\ref{acc-g}. For all of our models, we use the same Reddit development and test sets mentioned above.

For OntoNotes and EWT we use the entire corpora as datasets, without considering sub-genres within those resources, as most papers using them for training employ the entire corpus training set without internal distinctions. Although we recognize it would be interesting to analyze the contents of these data sets further, we leave that task open for future studies.

\subsection{Tagger}

For tagging, we choose a state-of-the-art neural sequence tagger, Flair~\cite{akbik-etal-2018-contextual} and retrain it on our splits. We used the sequence tagger model with the default 256 hidden unit bi-directional LSTM and trained with contextualized pre-trained Flair embeddings and uncontextualized pre-trained character embeddings, then evaluated performance by accuracy per token and also full-sentence accuracy (proportion of perfectly tagged sentences), since ``a single bad mistake in a sentence can greatly throw off the usefulness of a tagger to downstream tasks such as dependency parsing''~\cite{manning2011part}.

Finally, we use an ensemble approach to combine results from multiple models and study how much each of these sources contributes to the results of the ensemble model~\footnote{Setup details are available at~\url{https://github.com/shabnam-b/reddit-pos-ensemble.git}}. We use all the retrained Flair models on single genres except \textit{Reddit}, and then make predictions on the Reddit training set. We then use these predictions as training features for our ensemble model, which uses XGBoost as a meta-learner. Based on the analysis described in section~\ref{results} and to help the model better distinguish between NN and NNP, we also incorporate three other features to help the ensemble classifier; for each token, we check if 1) the token itself, 2) the lower-cased version of the token and 3) the token starting with a capital letter, exists in a knowledge base taken from \cite{ZeldesZhang2016} and add any entity types (e.g. Person, Organization etc.) which this token might have as n-hot encoded features. We then evaluate the classifier on the Reddit test set and perform an ablation study by removing the predictions of each genre and observing the changes in the accuracy.

\section{Results and Analysis}\label{results}

\begin{table*}[t]
\centering \fontsize{8.75}{12}\selectfont

\begin{tabular}{l|cccccc}
 & \textbf{Reddit} & \textbf{Multiple Genres}& \textbf{Multiple Genres w/o Reddit} & \textbf{Academic}& \textbf{Bio}& \textbf{Fiction}
 
\\
Training Set Size (Tokens) & 5,727 & 107,004& 101,277 & 11,868 & 12,562 & 12,843 \\
\hline
Per Token & 93.53 & \textbf{95.89}& 95.72&91.81 &91.77 &93.29  \\
Full-sentence & 36.08 & \textbf{53.16} & 49.37& 29.75& 25.32& 37.97\\

\\

& \textbf{Interview}& \textbf{News} & \textbf{Voyage}& \textbf{Whow} & \textbf{OntoNotes} & \textbf{EWT}\\
Training Set Size (Tokens) & 18,037 & 14,092 & 14,955& 16,920 & 2,442,000 &204,609\\
\hline
Per Token & 94.23 & 93.26 & 92.48& 92.95 &93.73&94.81\\
Full-sentence & 39.87 & 31.01& 30.38&31.01&41.14&48.10\\
\hline
\end{tabular}

\caption{\label{acc-g}
Accuracy scores calculated for tokens and full-sentences when trained on different genres individually and tested on Reddit.
}
\end{table*}

\begin{figure}[t]
\centerline{\includegraphics[scale=0.43]{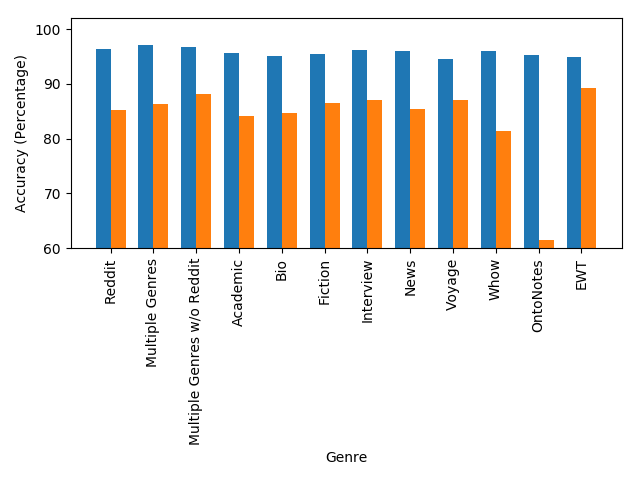}}
\caption{Accuracy on \textit{Known} and \textit{Unknown} tokens per genre; blue bars (left) correspond to accuracy of Known tokens and orange bars (right) correspond to accuracy of Unknown tokens.}
\label{fig1}
\end{figure}

\begin{table*}[t]
\centering \fontsize{8.75}{12}\selectfont

\begin{tabular}{l|l}
\textbf{Model} & \textbf{Example}\\
\hline

Reddit & \textit{\textcolor{red}{b./:} Using these to release \textcolor{red}{Boo/NN} into "The \textcolor{red}{Wild/NN}"} \\
&\textit{\textcolor{red}{love/VB} when I see people/places from Austin on FN :)}\\
 
Multiple Genres & \textit{\textcolor{red}{b./FW} Using these to release \textcolor{red}{Boo/NN} into "The \textcolor{red}{Wild/NN}"} \\
&\textit{\textcolor{red}{love/VB} when I see people/places from Austin on FN \textcolor{red}{:)/-RRB-}}\\

Multiple Genres w/o Reddit & \textit{\textcolor{red}{b./FW} Using these to release \textcolor{red}{Boo/NN} into "The Wild"} \\
&\textit{\textcolor{red}{love/VB} when I see people/places from Austin on FN \textcolor{red}{:)/:}}\\

Academic & \textit{\textcolor{red}{b./NN} Using these to release Boo into \textcolor{red}{"/DT} The Wild\textcolor{red}{"/CC}} \\
&\textit{\textcolor{red}{love/NN} when I see people/places from Austin on FN \textcolor{red}{:)/:}}\\

Bio & \textit{\textcolor{red}{b./FW} Using these to release Boo into "The Wild"} \\
&\textit{\textcolor{red}{love/NN} when I see people \textcolor{red}{//CC} places from Austin on FN \textcolor{red}{:)/:}}\\

Fiction & \textit{\textcolor{red}{b./``} Using these to release Boo into \textcolor{red}{"/NNP} The Wild"} \\
&\textit{\textcolor{red}{love/NN} when I see people \textcolor{red}{//:} places from Austin on FN \textcolor{red}{:)/:}}\\

Interview & \textit{\textcolor{red}{b./:} Using these to release Boo into "\textcolor{red}{The/NNP} Wild"} \\
&\textit{\textcolor{red}{love/VB} when I see people \textcolor{red}{//CC} places from Austin on FN \textcolor{red}{:)/:}}\\

News & \textit{\textcolor{red}{b./NNP} Using these to release \textcolor{red}{Boo/NN} into "The Wild"} \\
&\textit{\textcolor{red}{love/NN} when I see people/places from Austin on FN \textcolor{red}{:)/:}}\\

Voyage & \textit{\textcolor{red}{b./RB} Using these to release \textcolor{red}{Boo/NN} into "The Wild"} \\
&\textit{\textcolor{red}{love/VB} when I see people/places from Austin on FN \textcolor{red}{:)/:}}\\
 
Whow& \textit{\textcolor{red}{b./:} Using these to release \textcolor{red}{Boo/NN} into "The \textcolor{red}{Wild/NN}"} \\
&\textit{\textcolor{red}{love/VB} when I see people/places from Austin on FN \textcolor{red}{:)/:}}\\

OntoNotes & \textit{\textcolor{red}{b./NN} Using these to release Boo into "The Wild"} \\
&\textit{\textcolor{red}{love/VB} when I see people/places from Austin on FN \textcolor{red}{:)/.}}\\

EWT& \textit{\textcolor{red}{b./RB} Using these to release \textcolor{red}{Boo/NN} into "The Wild"} \\
&\textit{\textcolor{red}{love/VB} when I see people\textcolor{red}{//,}places from Austin on FN :)}\\

\hline
\end{tabular}

\caption{\label{examples}
Errors made by different models on two example sentences from Reddit posts.
}
\end{table*}

\begin{table}[t]
\centering \fontsize{8.75}{12}\selectfont
\begin{tabular}{l|c|c}
\textbf{Model} & \textbf{Per Token}&\textbf{Full-sentence}\\
\hline
StanfordNLP &94.81&46.84\\
TreeTagger&92.08&28.48\\
Ensemble & 95.99 & 53.80 \\
\ \ \ - (Academic) & 95.92 & 53.80 \\
\ \ \ - (Bio) & 95.89 &53.16\\
\ \ \ - (Fiction) & 95.95 & 54.43 \\
\ \ \ - (Interview) & \textbf{96.12} & \textbf{56.96} \\
\ \ \ - (News) & 95.89 &53.80 \\
\ \ \ - (Voyage) &95.92 &55.06 \\
\ \ \ - (Whow) & 95.82 &51.90\\
\ \ \ - (OntoNotes) & 95.55 & 50.00 \\
\ \ \ - (EWT) & 95.62 & 50.63 \\

\end{tabular}

\caption{\label{ens}
Accuracy score of StanfordNLP, TreeTagger and ensemble XGBoost when using the prediction of all trained models, and when each model is removed.
}
\end{table}

\begin{figure}[ht]
\centerline{\includegraphics[scale=0.43]{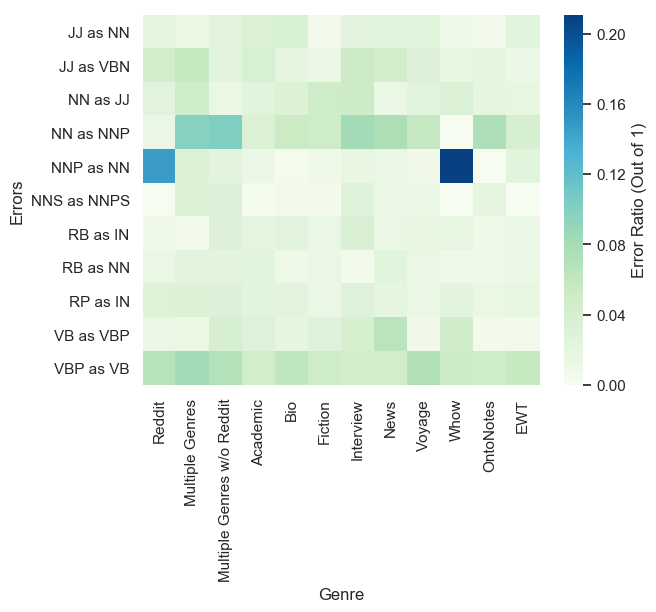}}
\caption{Most common prediction errors on the test set across models trained on different genres. Y-axis labeling: \textit{A as B} means A incorrectly tagged as B. }
\label{errors}
\end{figure}

The results of our experiments are shown in Table~\ref{acc-g}. As expected, the highest per token and also full-sentence accuracy belong to the model trained on multiple genres since there is more training data available and we are also including in-domain Reddit data in training. The model trained on multiple genres without Reddit gets only slightly lower accuracy per token, however, we observe a 3.2\% drop on full-sentence accuracy. Even though the Reddit model has the smallest amount of data for training (less than half of most other genres, due to the held out dev and test sets), it performs better than almost all models trained on different genres, which shows the high importance of even a small amount of in-domain data. Interestingly, the model trained on interviews works slightly better than the model trained on Reddit, probably because interviews published online are the most similar to the largely first and second person interactions found in Reddit forum discussions, and because the interview dataset is substantially larger than the Reddit training data.

In Table~\ref{examples}, we can compare errors made by different models on the same sentences. None of the models can predict correct POS tags for the whole sentences. Surprisingly, even though \textit{Multiple Genres} has more data than \textit{Multiple Genres w/o Reddit} including in-domain data, it performs worse in the first sentence; it cannot predict the tag NNP for the token `Wild'.

The most common error among all of the models is mistagging the token `b.' in the first sentence, which is indicating an item of a list and should get the tag LS. The second most common mistake seems to be the emoticon \textcolor{blue}{:)} in the second sentence. Reddit and EWT are the only models predicting correct tags for this token. `love' has the gold label VBP but it is predicted as VB or NN by different models, due to the low frequency of subjectless sentences, which resemble imperatives or fragments if the missing `I' is not recognized. NNP tokens such as `Boo' or `Wild'  are incorrectly tagged as NN by many of the models, mirroring findings on NNP tagging problems in previous studies.

We also look at the accuracy of models on \textit{Unknown} tokens (not seen during training) and \textit{Known} tokens separately. Figure~\ref{fig1} shows these results. Except for models trained on Academic, Bio, Whow and OntoNotes data, all other models perform better than the Reddit model on Unknown tokens, but this again could be the result of Reddit having a very small training set compared to other genres.

To further analyze the results, we looked at misclassifications which were common among multiple genres. The results are shown in Figure~\ref{errors}. The most common errors across all genres are VBP predicted as VB and NN predicted as NNP. The latter can stem from looser capitalization distinctions online, while the former can result when subject pronouns are dropped in informal English (e.g. `want to come?' or `need this right now'). Comparing \textit{Multiple Genres} and \textit{Multiple Genres w/o Reddit}, we can observe that adding the Reddit data results in more accurate RB, RP, and VB tagging. We can also observe that a huge proportion of the \textit{Whow} model's errors belongs to mistagging NNP as NN, which is probably the result of fewer proper nouns appearing in Wikihow articles since they are sets of instructions for various tasks.

Furthermore, we manually observed 50 of the errors that the \textit{Multiple Genres w/o Reddit} model made. The most common errors were 1) \textit{Emoticons:} Emoticons such as \textcolor{blue}{:)} , \textcolor{blue}{:(} or others which are gold labeled as SYM are labeled with different tags such as \textcolor{blue}{"}, \textcolor{blue}{:} or even NNP in cases where they contain an alphabetical character such as in \textcolor{blue}{D:$>$}.
2) \textit{Interjections} and mostly swear words appear in social media text more than other genres, as well as phonetic elongation or representations with repeated characters such as `NANANANA', which do not appear in formal written text, but are common among users in social media \cite{SanguinettiBoscoCassidyEtAl2020}. Some of these tokens were tagged as NNP instead of gold standard UH. Some other errors were 3) \textit{Proper nouns not starting with a capital letter} (e.g. bobby/NN), 4) \textit{Foreign words} (e.g. etcetera/NNP) and 5) \textit{Abbreviations} (e.g. BTW/NNP).

Finally, in order to harness the increased stability offered by consulting multiple models and different features, Table~\ref{ens} shows the results of the ensemble model described in Section~\ref{app}. Except for Interview, all models positively contribute to the overall accuracy. Only the Interview model's removal from the ensemble improves upon the results in Table~\ref{acc-g} both in terms of per token accuracy and full sentence accuracy, which suggests that, at least for the test set at hand, other genres combined do a better job of predicting correct tags, despite the usefulness of interviews in a single genre model. The final model without `interview' thus represents our best results and a new state-of-the-art score on Reddit tagging using the GUM benchmark, with 96.12\% accuracy despite not including any Reddit data in training the features for the meta-classifier. Furthermore, we compare these results with two pretrained off-the-shelf taggers: TreeTagger trained on Penn treebank and StanfordNLP~\cite{qi2018universal} trained on GUM. We also looked at the effect of removing the named entity features on the results; without the named entities, the best model's accuracy (Ensemble-Interview) drops to 95.89\% per token and 55.06\% for full-sentence. Comparing these numbers to the best single model in Table~\ref{acc-g}, the ensemble approach without any extra features is resulting in the same token accuracy, but we get almost 2\% increase in full-sentence accuracy.

\section{Conclusion}
In this work, we looked at the effect of genre on Reddit POS tagging. We analyzed the results of the same tagger, trained on 10 different sources with multiple genres, including Reddit itself, and also trained on the combination of all genres. We observed that within single genre models, the Interview model has the highest accuracy on Reddit, which might be the result of comparatively much training data (Interview has somewhat more tokens than the other single genre datasets), or the nature of some of the Reddit documents which are back and forth conversations between different users and is similar to the nature of interviews. However, in combination with other models in the ensemble approach, removing the Interview model seemed to increase the performance slightly, and OntoNotes predictions seem to have the most positive contributions to the accuracy of the ensemble model, possibly because of the wide coverage of rare items resulting from the large corpus size.

Finally, the results of our error analysis were in line with prior studies on Web text-types other than Reddit. The most important problem in this task using deep learning models with word/character embeddings is when we have unseen data in the test set; this unseen data could be in the form of creative emoticons, repeated characters in a word, abbreviations, etc. To improve the performance on user-generated content in domains such as social media, we either need to collect sufficient in-domain data to train a genre-specific model, or find other ways of addressing unseen tokens such as using lexical resources or doing specific preprocessing to normalize the tokens before testing. The present paper demonstrates that, in the absence of substantial amounts of in-domain data, ensembling the outputs of multiple tagging models with different training datasets can lead to very good results, in this case giving a new SOA score of 96.12\% token accuracy on the Reddit data. At the same time, full-sentence accuracy remains below 57\%, suggesting that there is still room for substantial improvements.

% \nocite{*}
\section{Bibliographical References}\label{reference}
%\label{main:ref}

\bibliographystyle{lrec}
\bibliography{mybib}

\section{Language Resource References}
\label{lr:ref}
\bibliographystylelanguageresource{lrec}
\bibliographylanguageresource{languageresource}

\end{document}